\documentclass[1p]{elsarticle}

\usepackage{hyperref}

\usepackage{graphicx}
\usepackage{subfigure}
\usepackage{amsmath,amssymb}
\usepackage{multirow}
\usepackage{algorithm}
\usepackage[noend]{algorithmic}

\usepackage[utf8]{inputenc}

\journal{Pattern Recognition}

\begin{document}
	
	\begin{frontmatter}
		
		\title{Discriminative feature generation for classification of imbalanced data}
		
		\author[KISTEurope,TUKaiserslautern]{Sungho Suh}
		\ead[url]{https://github.com/opensuh/DFG/}
		\author[TUKaiserslautern,DFKI]{Paul Lukowicz}
		\author[KISTEurope]{Yong Oh Lee\corref{mycorrespondingauthor}}
		\cortext[mycorrespondingauthor]{Corresponding author}
		\ead{yongoh.lee@kist-europe.de}

		\address[KISTEurope]{Smart Convergence Group, Korea Institute of Science and Technology Europe Forschungsgesellschaft mbH, 66123 Saarbrücken, Germany}
		\address[TUKaiserslautern]{Department of Computer Science, TU Kaiserslautern, 67663 Kaiserslautern, Germany}
		\address[DFKI]{German Research Center for Artificial Intelligence (DFKI), 67663 Kaiserslautern, Germany}

		\begin{abstract}
			The data imbalance problem is a frequent bottleneck in the classification performance of neural networks. In this paper, we propose a novel supervised discriminative feature generation (DFG) method for a minority class dataset. DFG is based on the modified structure of a generative adversarial network consisting of four independent networks: generator, discriminator, feature extractor, and classifier. To augment the selected discriminative features of the minority class data by adopting an attention mechanism, the generator for the class-imbalanced target task is trained, and the feature extractor and classifier are regularized using the pre-trained features from a large source data. The experimental results show that the DFG generator enhances the augmentation of the label-preserved and diverse features, and the classification results are significantly improved on the target task. The feature generation model can contribute greatly to the development of data augmentation methods through discriminative feature generation and supervised attention methods.
		\end{abstract}
		
		
	\end{frontmatter}
	
	
	\section{Introduction}
	\label{introduction}
	
	Deep learning has achieved significant improvements in various tasks in computer vision applications using open image datasets containing a large amount of data. However, the acquisition of large datasets is a challenge in real-world applications, especially if they pertain to new areas in deep learning. For training deep neural networks, the quantity of data is as important as the quality because training with only a small dataset often results in degradation of the neural network performance or introduces an over-fitting problem. Attempts have been made to establish a large training dataset in various fields; however, the annotation of a dataset still remains expensive, laborious, and time-consuming. Furthermore, the distribution of classes in the dataset is often imbalanced. Training a network under class-imbalanced conditions can produce a detrimental effect on neural network performance and a biased classification result \cite{japkowicz2002class, xie2007effect, buda2018systematic}. 
	
	Data augmentation enhances the size and quality of the training dataset. Techniques range from simple data augmentation, such as flip, shift, and rotation, to deep generative models based on generative adversarial networks (GAN) \cite{goodfellow2014generative}, where a small training dataset is augmented at the input data level. Because GANs can approximate the distribution of the real input data and generate realistic samples from a generative model, many recent studies have shown that augmenting a small training dataset using a GAN can improve the classification performance in real applications \cite{lee2017application, douzas2018effective, suh2019generative, huang2017stacked, guo2019discriminative, suh2020two}. However, such data augmentation at input level showed the limited improvement in the performance, because balancing the data distribution have week relation to the enhancement of feature extraction in the minority dataset. Also, the state-of-art GAN has the limited generating capability on large-scale images \cite{lucic2018gans, brock2019large}. When the size of input image is large, these augmentation method is not applicable.
	
	Unlike data augmentation, adversarial feature augmentation generates domain-invariant features, increasing the size of the minority classes in the feature space without considering the modality of the input data \cite{li2018delta}. This augmentation was adopted in supervised learning, especially for the class imbalance problem\cite{zhang2019feature}. Although the feature augmentation method performs better than the unsupervised domain adaptation method by generating domain-invariant and modality-free features, the generated feature maps from the outer fully connected layer do not have sufficient feature dimensions to represent the data distribution with a small amount of minority class data. In other words, generating features in convolutional layers has a greater capacity to improve classification performance in the case of the class-imbalanced conditions.
	
	Taking these problems into consideration, we propose a novel discriminative feature generation (DFG) method using attention maps in the feature space. The proposed method is a combination of transfer learning and adversarial feature augmentation to complement their drawbacks. The baseline is transfer learning with regularizations, and the features are augmented using a GAN with the weight of the activation level of the feature maps in each class. 

	The framework is based on a modified GAN structure containing an independent classifier to improve the neural network performance. We extend the adversarial feature augmentation \cite{volpi2018adversarial} in the following three manners: (1) Our GAN structure is composed of four independent networks: a feature generator, feature discriminator, feature extractor, and feature classifier. The more constituent networks enable the network to be trained for generating more effective features, especially for small sample data, in the feature space. (2) Similar to DELTA \cite{li2018delta}, we deploy regularizations to the feature extractor and the feature classifier for transfer learning to select the discriminative features from the outer layer outputs and transfer pre-trained knowledge from the large-scale source dataset. (3) In the training phase, we employ a feature generation method in a small-sized target dataset, which augments the selected discriminative features by adopting the attention mechanisms for each class label through supervised learning. 
	
	Our main contribution is a novel feature generation and augmentation using the proposed GAN structure to unravel the data imbalanced problem and improve neural network performance. In the proposed GAN structure, a feature extractor and a feature classifier are included to train together with a feature generator and a feature discriminator. For the generation of meaningful features for classification of small-sized target data, transfer learning with regularization and class-wise attention is adopted. We evaluate the proposed method with various pairs of source and target dataset to show general applicability to classification in the class-imbalanced conditions.
	
	The remainder of this paper is organized as follows. In Section \ref{relatedwork}, related works are summarized. The framework and training procedure are presented in Section \ref{method}. In Section \ref{experimentalresult}, the benchmark datasets for verification and the experimental results are detailed. Section \ref{conclusions} concludes the paper and provides scope for future research.

	\section{Related Work}
	\label{relatedwork}
	
	GAN \cite{goodfellow2014generative} is a generative model based on a min-max game theory scenario that pits two networks against one another. A generator network, $G$, competes against a discriminator network, $D$, that distinguishes between samples generated from $G$ and samples from the training data. GAN can generate synthetic data close to the original; however, the training process of GAN has the instability of loss function convergence and the problem of mode collapse \cite{goodfellow2016nips}. To avoid such problems, the Wasserstein GAN (WGAN) \cite{arjovsky2017WGAN} and the WGAN with a gradient penalty (WGAN-GP) \cite{gulrajani2017improved} using the Wasserstein-K distance as the loss function were proposed. Recently, GAN has been extended in several ways to control the generated properties \cite{mirza2014conditional, odena2017conditional, lee2019controllable} and its utilization, such as data augmentation \cite{douzas2018effective, cenggoro2018deep, suh2019generative}. 

	Recently, GAN-based data augmentation methods were proposed to further improve the creation of augmented synthetic training data. Huang et al. \cite{huang2017stacked} proposed stacked GAN (SGAN), which is trained to invert the hierarchical representations of a bottom-up discriminative network. Guo et al. \cite{guo2019discriminative} proposed a discriminative variational autoencoding adversarial network, which learns a latent two-component mixture distributor and alleviates the class imbalance for deep imbalanced learning. Cui et al. \cite{cui2019class} proposed a class-balanced loss for long-tailed distributions. The class-balanced loss re-weights losses inversely with the effective number of samples per class. Although the generated synthetic data can balance the distribution between the classes, the classification performance is limited due to no guarantee enhancing the feature extraction ability of the classifier, when it is an independent network of the GAN. Even if GAN is trained with the classifier together like \cite{suh2020cegan}, GAN has the limitation to generate real-looking synthetic data on large-scale images. Not-qualified synthetic data is not able to improve the performance of the classification significantly.
	
	Unlike GAN-based data augmentation at the input data level, Volpi et al. \cite{volpi2018adversarial} proposed augmentation in the feature space. A feature extractor is trained with the source dataset under supervised learning, and then a feature generator for the unlabeled target dataset is trained in the convolutional GAN (CGAN) framework \cite{mirza2014conditional} against the feature extractor. Zhang et al. \cite{zhang2019feature} developed a more general feature generation framework for imbalanced classification, inspired by the adversarial feature augmentation approach. These methods can generate domain-invariant features without considering the modality of the data. Consequently, they achieved better performance than unsupervised domain adaptation methods. However, the improvement of the classification under the class-imbalanced condition is still not significant. The dimension of the features in the feature extractor is considerably lower than the input data. Such a low dimension may not be sufficient to present the data distribution of a small amount of minority class data. The generated features should be domain-invariant and modality-free.

	Transfer learning is a machine learning method that transfers knowledge learned in a source task to a target task \cite{caruana1997multitask, pan2009survey, donahue2014decaf}. The weights of deep neural networks are first pre-trained on a large-scale dataset, which is called the source task, and then fine-tuned using the data from the target task with a small amount of data \cite{pan2009survey}. In the case of fine-tuning in convolutional neural networks, the weights in the first few convolution layers are fixed by pre-training. The last convolution layers are fine-tuned by the target task. During fine-tuning, the parameters of the target model can be driven far away from the pre-trained parameter values, leading to the incorporation of information relevant to the targeted problem and overfitting to the target task, so-called catastrophic forgetting \cite{kirkpatrick2017overcoming}. This simple method cannot guarantee good performance because it may in many cases burden the network of the target task with irrelevant information. Yosinki et al. \cite{yosinski2014transferable} quantified the transferability of features from each layer to learn transferable representations. For constraining catastrophic forgetting in inductive transfer learning, L$^2$-norm regularization was proposed in \cite{li2018explicit}. The key concept of L$^2$-SP is ``starting point as reference'' optimization, which tries to drive weight parameters to pre-trained values by regularizing the distance between the parameters of source and target tasks.
	
	Instead of regularizing the weight parameters, deep learning transfer using a feature map with attention (DELTA), which is a regularized transfer learning framework, was proposed by Li et al. \cite{li2018delta}. Inspired by knowledge distillation for model compression \cite{hinton2015distilling, Zagoruyko2017AT, yim2017gift}, DELTA employs the ideas of ``inactivated channel re-usage'' and feature map regularization with attention. These regularization approaches achieved significant improvement and alleviated the catastrophic forgetting problem by drawing weight parameters close to pre-trained values or aligning transferable channels in feature maps. It constrains the difference between the feature maps generated by the convolution layers of the source and target networks with attention. DELTA selects the discriminative features from the outer layer outputs using a supervised attention mechanism. To further improve the performance of the classification with a small training dataset or class-imbalanced dataset, a DFG method using attention maps in the feature space is proposed in this paper.
	
	\section{Discriminative Feature Generation}
	\label{method}
	
	\begin{figure}
		\centering
		\includegraphics[width=0.95\columnwidth]{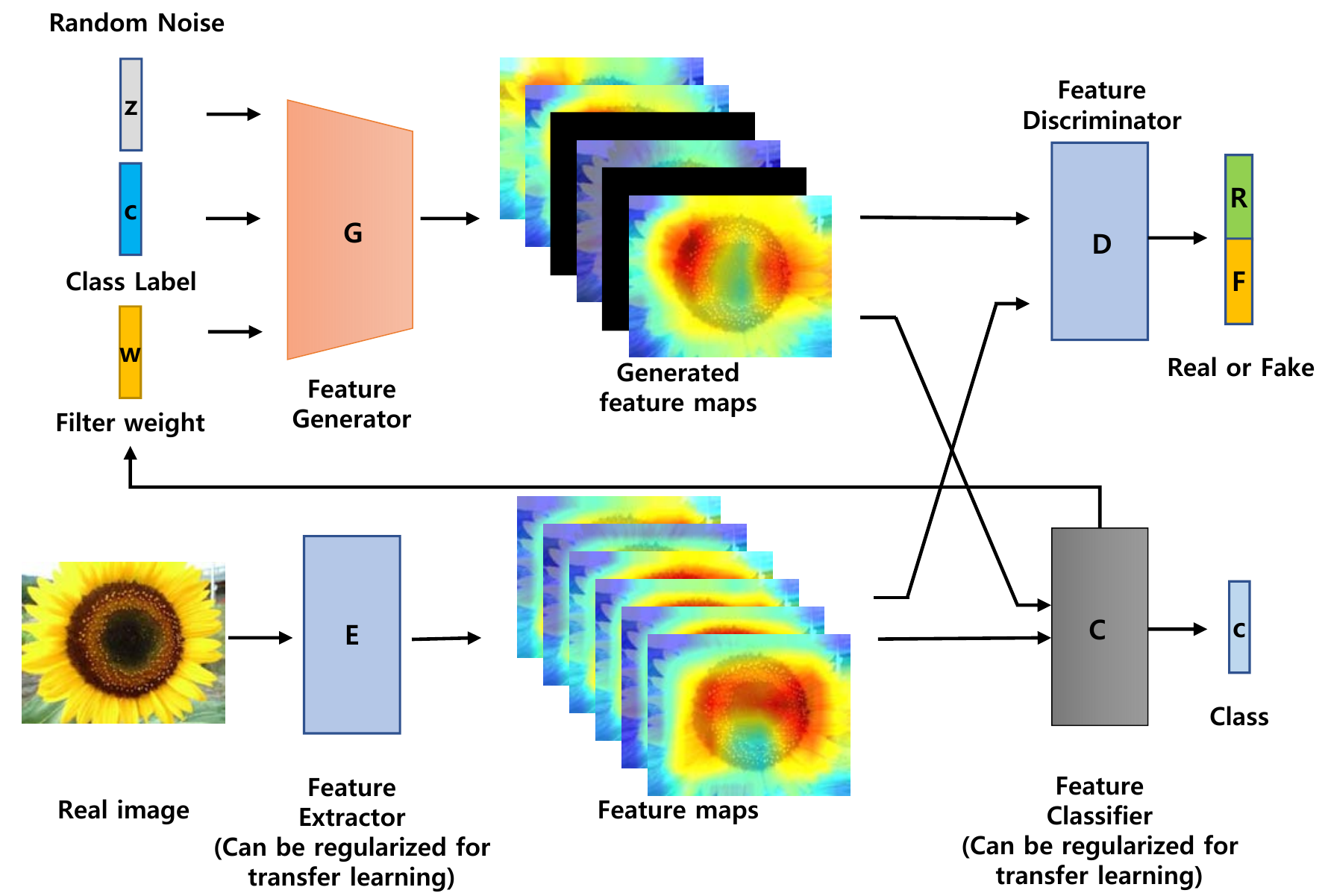}
		\caption{Structure of the proposed model.}
		\label{trainingprocedure}
	\end{figure}	
	
	Our goal is to generate discriminative features to improve the performance of the neural network on a small number of training data and the class-imbalanced dataset. 
	
	Unlike CGAN and ACGAN\cite{mirza2014conditional, odena2017conditional}, the structure of the feature classifier to the independent network is employed in the proposed GAN structure. The proposed GAN structure is composed of four independent networks: a feature generator, feature discriminator, feature extractor, and feature classifier. The structure of the proposed GAN model is shown in Figure \ref{trainingprocedure}. The feature extractor represents a first convolution layer or block in a neural network classifier, and the feature classifier represents the rest of the convolution layers and fully connected layers after the feature extractor in the neural network classifier. The feature extractor and classifier are regularized with transferred weights from the source model, similar to Li et al. \cite{li2018delta}. We deploy the knowledge distillation technique to the training procedure of GAN and train the feature generator with supervised attention models to generate discriminative features.  
	
	\subsection{Training GAN model with four independent networks}
	For the stability of the training procedure and the quality of the generated data, we deploy the objective formulation of WGAN-GP for the feature discriminator and the feature generator. The loss functions of the feature discriminator and the feature generator are denoted by $\mathop{\mathbb{L}_D}$ and $\mathop{\mathbb{L}_G}$, respectively.
	
	\begin{equation}
	\label{eq:dloss}
	\mathop{\mathbb{L}_D}(x,z,\hat{y};\theta_D) = - \mathop{\mathbb{E}}_{(z,\hat{y})\sim (P_z, \hat{Y})}[D(G(z,\hat{y}))] + \mathop{\mathbb{E}}_{x\sim X}[D(E(x))] +  \lambda \mathop{\mathbb{E}}_{\hat{x}\sim P_{\hat{x}}}[(\Arrowvert \nabla_{\hat{x}} D(\hat{x}) \Arrowvert_2 -1 )^2],
	\end{equation}
	\begin{equation}
	\label{eq:gloss}
	\mathop{\mathbb{L}_G}(z,\hat{y};\theta_G) = -\mathop{\mathbb{E}}_{(z,\hat{y})\sim (P_z, \hat{Y})}[D(G(z,\hat{y}))],
	\end{equation}
	where $z$ is the noise vector sampled from uniform distribution $P_z$, and feature generator $G$ generates synthetic feature $G(z)$. $\lambda$ is the penalty coefficient, and $P_{\hat{x}}$ is the uniform sampling along straight lines between pairs of points from the real data distribution, $P_r$, and the generated data distribution. $\theta_D$ and $\theta_G$ are the parameters of the feature discriminator and the feature generator, respectively. Feature discriminator $D$ is trained to minimize $\mathop{\mathbb{L}_D}$ to distinguish between real and generated features. Generator $G$ is trained to minimize $\mathop{\mathbb{L}_G}$.
	
	To generate the distribution of features similar to the real features from the feature extractor, a total loss function of the feature classifier contains a loss function of the feature classifier on the features concatenating the real and generated features as well. The objective functions of the feature extractor and classifier are represented as follows: 
	
	\begin{equation}
	\label{eq:eloss}
	\mathop{\mathbb{L}_E}(x,y;\theta_E) = \mathop{\mathbb{E}}_{(x,y)\sim (X, Y)} [-y\log C(E(x))],
	\end{equation}
	\begin{subequations}\label{eq:closs}
		\begin{equation}\tag{\ref{eq:closs}}
		\begin{split}
		\mathop{\mathbb{L}_C}(x,y,z,\hat{y};\theta_C) = &\alpha \mathop{\mathbb{L}^r_C}(x,y;\theta_C) + \beta \mathop{\mathbb{L}^r_C}(x,y;\theta_C) + \gamma L^c_C\\ 
		&(\alpha + \beta + \gamma = 1).
		\end{split}
		\end{equation}
		In (\ref{eq:closs}), the loss functions of $L^r_C$, $L^g_C$, and $L^c_C$ are defined as
		\begin{equation}
		\begin{split}
		\mathop{\mathbb{L}^r_C}(x,y;\theta_C) = \mathop{\mathbb{E}}_{(x,y)\sim (X, Y)} [-y\log C(E(x))],
		\end{split}
		\label{eq:clossa}
		\end{equation}
		\begin{equation}
		\begin{split}
		\mathop{\mathbb{L}^g_C}(z,\hat{y};\theta_C) = \mathop{\mathbb{E}}_{(z,\hat{y})\sim (p_z(z),\hat{Y})} [-\hat{y}\log C(G(z,\hat{y}))],
		\end{split}
		\label{eq:clossb}
		\end{equation}
		\begin{equation}
		\begin{split}
		\mathop{\mathbb{L}^c_C}(x,y,z,\hat{y};\theta_C,\theta_G) = &\mathop{\mathbb{E}}_{(\tilde{x}, \tilde{y}) \sim (\tilde{X}, \tilde{Y})} [-\tilde{y} \log C(\tilde{x})],\\ 
		&\text{where}~\tilde{x} = E(x) \oplus G(z,\hat{y}),~ \tilde{y} = y \oplus \hat{y} 
		\end{split}
		\label{eq:clossc}
		\end{equation}
	\end{subequations}
	where $\mathop{\mathbb{L}_E}$ and $\mathop{\mathbb{L}_C}$ are the loss functions of the feature extractor and feature classifier, respectively. $\theta_E$ and $\theta_C$ are the parameters $E$ and $C$, respectively.  $\oplus$ denotes a concatenation operation. $\mathop{\mathbb{L}^c_C}$ is the loss function of the feature classifier on the feature maps concatenated with the real and generated features. $\alpha$, $\beta$, and $\gamma$ are hyperparameters that control the importance of the classification for the real and generated data. In $\mathop{\mathbb{L}_E}$ and $\mathop{\mathbb{L}^r_C}$, the regularization term, which characterizes the differences between the source and target network for transfer learning, can be added.
	
	Feature generator $G$ (to minimize (\ref{eq:gloss}) and (\ref{eq:clossc})) and feature classifier $C$ (to minimize (\ref{eq:closs}) and (\ref{eq:clossa})) are trained simultaneously, whereas feature discriminator $D$ and feature extractor $E$ are trained to minimize (\ref{eq:dloss}) and (\ref{eq:eloss}), respectively. The reason why feature classifier $C$ is optimized not only with the real data is to prevent the classifier from overfitting to the real data because the performance of the classifier with the real data is not sufficient to improve the performance of the classification under the class-imbalanced condition. 
	
	In the training procedure, we define two training parameters. The generator learning parameter controls the ratio between $\mathop{\mathbb{L}_G}$ and $\mathop{\mathbb{L}^g_C}$ while optimizing the generator parameter, $\theta_G$. The classifier learning parameter is used to control the balance of how much the classifier is trained from the feature extracted from the real data $\mathop{\mathbb{L}^r_C}$ or from the real features combined with the generated features, $\mathop{\mathbb{L}_C}$. The training details of the proposed method using (\ref{eq:dloss}), (\ref{eq:gloss}), (\ref{eq:eloss}), and (\ref{eq:closs}) with these parameters are summarized in Algorithm \ref{Algopseudocode}.
	
	\subsection{Discriminative feature generation by using the supervised attention model}
	\begin{algorithm}
		\caption{Training procedure of the proposed DFG method. We use the default values of $\lambda$ = 10, $n_D$ = 5, $n_{C1} = 2$, $n_{C2} = 10, and n_W = 2000$}\label{Algopseudocode}
		\begin{algorithmic}[1]
			\REQUIRE Batch size $m$, learning rate $\eta$, hyperparameter for weight sum $\rho$, hyperparameters $\alpha$, $\beta$, and $\gamma$, and a threshold value for filter weight $\delta$.
			\STATE \textbf{Initialize:} $\theta_E$, $\theta_C$ from pre-trained source networks $\theta_{E_S}$, $\theta_{C_S}$.
			\STATE $W_j(E_S;\theta_{E_S}) \gets $ softmax($-y\log C_S(E_{\theta_{E_S}^{\backslash j}}(x)) + y\log C_S(E_{\theta_{E_S}}(x))$) 
			\FOR{$k=1$,..., $n_l$ number of class labels}
			\STATE $W_j^*(E_S, y=k;\theta_{E_S}) \gets W_j$ (if $W_j > \delta / n_l$), $W_j^*(E_S, y=k;\theta_{E_S}) \gets 0$ (else)
			\ENDFOR		
			\FOR{$step=1,...,$ number of training iteration}
			\FOR{$t=1,...,n_D$}
			\STATE Sample $\{x^{(i)}\}^m_{i=1}\sim P_r$ a batch from the real data.
			\STATE Sample $\{z^{(i)}\}^m_{i=1}\sim P_z$ a generated batch and labels $\{\hat{y}^{(i)}\}^m_{i=1}$.
			\STATE Update feature discriminator $D$ using Eq. (\ref{eq:dloss}): $\theta_D \gets \theta_D - \eta_D \nabla_{\theta_D} \mathop{\mathbb{L}_D}(x,z,\hat{y};\theta_D)$, 
			\ENDFOR
			\STATE Sample $\{z^{(i)}\}^m_{i=1}\sim P_z$ a generated batch and labels $\{\hat{y}^{(i)}\}^m_{i=1}$.
			\STATE Sample $\{x^{(i)}\}^m_{i=1}\sim P_r$ a batch from the real data and labels $\{y^{(i)}\}^m_{i=1}$.
			\STATE Update feature generator $G$ using Eq. (\ref{eq:gloss}): $\theta_G \gets \theta_G - \eta_G \nabla_{\theta_G} \mathop{\mathbb{L}_G}(z,\hat{y},W_j^*;\theta_G)$, 
			\IF{$step$ $\equiv 0 (\mod n_{C1})$}
			\STATE Concatenate real and generated features following Eq. (\ref{eq:clossc}): $\tilde{x}$, $\tilde{y}$, 
			\STATE Update $G$ using Eq. (\ref{eq:clossc}): $\theta_G \gets \theta_G - \eta_G\nabla_{\theta_G} \mathop{\mathbb{L}^c_C}(x,y,z,\hat{y},W_j^*;\theta_G)$, 
			\STATE Update feature extractor $E$ using Eq. (\ref{eq:eloss}): $\theta_E \gets \theta_E - \eta_E\nabla_{\theta_E} \mathop{\mathbb{L}_E}(x,y;\theta_E)$, 
			\STATE Update feature classifier $C$ using Eq. (\ref{eq:clossa}): $\theta_C \gets \theta_C - \eta_C\nabla_{\theta_C} \mathop{\mathbb{L}^r_C}(x,y;\theta_C)$, 
			\ENDIF
			\IF{$step$ $\equiv 0 (\mod n_{C2})$}
			\STATE Concatenate real and generated features following Eq. (\ref{eq:clossc}): $\tilde{x}$, $\tilde{y}$, 
			\STATE Update $C$ using Eq. (\ref{eq:closs}): $\theta_C \gets \theta_C - \eta_C\nabla_{\theta_C} \mathop{\mathbb{L}_C}(x,y,z,\hat{y},W_j^*;\theta_C)$, 
			\ENDIF
			\IF{iter $\equiv 0 (\mod n_W)$}
			\STATE Update filter weight using Eq.(\ref{eq:WeightingFeaturemaps}): $W_j(E;\theta_{E}) \gets $ softmax($-y\log C(E_{\theta_E^{\backslash j}}(x)) + y\log C(E_{\theta_E}(x))$). 
			\FOR{$k=1$,..., $n_l$ number of class labels}
			\STATE $W_j^*(E,y=k;\theta_E) \gets \rho W_j(E_S,y=k;\theta_{E_S}) + (1-\rho) W_j(E,y=k;\theta_E)$ Eq.(\ref{eq:classwisegenerator_attention_weightedsum}) (if $W_j(E;\theta_E) > \delta / n_l$),  
			\STATE $W_j^*(E,y=k;\theta_E) \gets 0$ (else)
			\ENDFOR	
			\ENDIF
			\ENDFOR
		\end{algorithmic}
	\end{algorithm}
	
	To generate discriminative features, we adopt supervised attention mechanisms for each class label. To obtain the weights for feature maps, we propose a supervised attention method adopted from \cite{li2018delta} for the generator network. Whereas the supervised attention method in \cite{li2018delta} is calculated by averaging the filter weight for each filter, we utilize the fact that the importance of each filter varies from class to class. We transform the filter weights for a single data into class-wise filter weights and deactivate channels with low filter weights. 
	
	In the supervised attention method from \cite{li2018delta}, the weights of the features are characterized by the performance loss when removing the convolutional filter for each feature from the feature extractor network. For a conv2d layer in the feature extractor and generator, the parameter form is a four-dimensional tensor with the shape of $(c_{i+1}, c_i, k_h, k_w)$, where $c_i$ denotes the number of channels of the $i$-th layer and $(k_h, k_w)$ represents the size of the kernel. At the last convolution layer of the feature extractor, the dimension of the output feature maps is $(c_{i+1}, h_{i+1}, w_{i+1})$, which is matched with the dimension of the output of the feature generator. We can measure significance through the performance reduction of the feature extractor and classifier when the filter in the last convolution layer is disabled in the feature extractor network. In other words, because it usually causes higher performance loss to remove a filter with a greater capacity for discrimination, we can improve the performance of the classification by generating feature maps focusing on channels with high filter weights. The filter weight is calculated through the feature extractor and classifier and is input into the feature generator. The feature generator generates weighted feature maps with deactivated feature maps using the filter weight. The deactivated feature maps are denoted in black in Figure \ref{trainingprocedure}. 
	
	The filter weight is expressed in (\ref{eq:WeightingFeaturemaps}), which is used to calculate the gap between the classification losses of the feature extractor and classifier on data $(x,y)$ with and without the $j$-th filter in the last convolution layer of the feature extraction network.
	
	\begin{equation}
	\label{eq:WeightingFeaturemaps}
	\begin{split}
	W_j(E,x,y;\theta_E) &= \text{softmax} [-y\log C(E_{\theta_E^{\backslash j}}(x)) + y\log C(E_{\theta_E}(x))], 
	\end{split}
	\end{equation}
	where $\theta_E^{\backslash j}$ denotes the modified parameter from $\theta_E$ with all elements of the $j$-th filter set to zero. 
	
	Because each class usually has different importance of the filter channel, the calculated filter weights for a single data point can be transformed into class-wise filter weights.
	\begin{equation}
	\label{eq:classwisegenerator_attention}
	\begin{split}
	W_j(E,c;\theta_E) = n_f\times \frac{W_j(E, x_i, y_i|y_i=c;\theta_E)}{n_c}
	\end{split}
	\end{equation}
	where $W_j(E,c;\theta_E)$ ($W\in \mathcal{R}^{n_c\times n_f}$) is a class-wise filter weight, $c$ is a class label, $n_c$ is the number of images in the class, and $n_f$ is the number of filters. By multiplying the number of filters $n_f$, we set the average value of the weights to 1. When transfer learning is used for the small training dataset and class imbalance problem, a filter weight on the source model can be obtained through feature extractor $E_S$ with parameter $\theta_{E_S}$ and classifier $C_S$ with parameter $\theta_{C_S}$ trained on the source dataset. Whereas the filter weight in \cite{li2018delta} was calculated on the source model only, the proposed method updates the filter weight on the target model. In this case, the filter weight can be expressed in the form of a weighted sum between the filter weight on the source dataset and the target dataset.
	
	\begin{equation}
	\label{eq:classwisegenerator_attention_weightedsum}
	\begin{split}
	W_j(E,c;\theta_E) = \rho W_j(E_S,c;\theta_{E_S}) + (1-\rho)W_j(E,c;\theta_E), 
	\end{split}
	\end{equation}
	
	By applying the weight to the filters of the last layer in the feature generator, the feature generator can generate discriminative features by following the supervised attention models. The operation of the last layer in the original feature generator is expressed as follows.
	\begin{equation}
	\label{eq:originalfeaturegenerator}
	\begin{split}
	f^l(z,\hat{y})=\text{norm}[\tanh(\text{TransConv}(f^{l-1}(z,\hat{y}),\theta_G^l)+b^l)]
	\end{split}
	\end{equation}
	Note that $f^l$ denotes the features of the $l$-th layer in the feature generator, $\text{norm}()$ performs batch normalization, $\text{TransConv}(f,\theta)$ is a transpose convolution that takes $f$ as the input with $\theta$ as a weight, and $b^l$ is a bias vector of the $l$-th layer. To generate discriminative features by using the class-wise weights of a filter, the operation can be rewritten as follows.
	\begin{equation}
	\label{eq:weightedfeaturegenerator}
	\begin{split}
	\hat{f}^l_j(z,\hat{y})=\text{norm}[W_j^*(E,c=\hat{y};\theta_E)\cdot \tanh(\text{TransConv}_j(f^{l-1}(z,\hat{y}),\theta_G^l)+b^l_j)]
	\end{split}
	\end{equation}
	
	$\hat{f}^l(z,\hat{y})$ is a discriminative feature generated by the supervised attention model. To maximize the difference in the weights and inactivate unimportant filter channels, we set the weight values that are smaller than a threshold, $\delta$, to zero and represent as $W_j^*$. $\delta$ is set to 0.95 of the average value of all filters for each class.

	\section{Experimental Results}
	\label{experimentalresult}
	\subsection{Datasets and architectures}
	To evaluate the proposed method, we used the following six benchmark datasets: Street View House Numbers (SVHN) \cite{netzer2011reading}, Fashion-MNIST (F-MNIST) \cite{xiao2017fashion}, STL-10 \cite{coates2011analysis}, CINIC-10 \cite{darlow2018cinic}, Caltech-256 \cite{griffin2007caltech}, and Food-101 \cite{bossard2014food}. As a classifier model, LeNet5 \cite{lecun1998gradient} for SVHN and F-MNIST, VGGNet-16 \cite{simonyan2014very} for STL-10 and CINIC-10, and ResNet-50 \cite{he2016deep} for Caltech-256 and Food-101 were employed. We used extended MNIST digits (EMNIST) \cite{cohen2017emnist}, CIFAR-10 \cite{krizhevsky2009learning}, and ImageNet \cite{deng2009imagenet} as the source domain for LeNet-5, VGG-16, and ResNet-50, respectively. The architecture of our feature discriminator and feature generator closely follow the deep convolutional GAN (DCGAN) architecture model \cite{radford2015unsupervised}, which is an extended model of the GAN that uses three transpose convolution layers in the generator and three convolution layers in the feature discriminator. A summary of the benchmarked dataset is presented in Table \ref{dataset}. \ref{networkarchitectures} contains additional information on benchmarks and hyperparameters. The source code to reproduce our experiments is available at \url{https://github.com/opensuh/DFG}.
	
	\begin{table}[t]
		\caption{Summary of the benchmark datasets.}
		\label{dataset}
		\begin{center}
			\begin{small}
				\begin{sc}
					\begin{tabular}{ccc}
						\hline
						Dataset & Classifier & \# of training dataset \\
						\hline
						SVHN & LeNet5 & 73237 \\
						F-MNIST & LeNet5 & 60000 \\
						STL-10 & VGGNet-16 & 5000\\
						CINIC-10 & VGGNet-16 & 18000\\
						Caltech-256 & ImageNet & 30607\\
						Food-101 & ImageNet & 75750\\
						\hline
					\end{tabular}
				\end{sc}
			\end{small}
		\end{center}
	\end{table}
	
	To evaluate the proposed method under the class-imbalanced condition, we adopt the step imbalance type described in \cite{buda2018systematic}. The step imbalance assumes that the classes are divided into minority and majority groups. Then, classes in the same group have the same number of data points, making a step in the data distribution plot. In this evaluation, we set the number of majority classes as two and eight to evaluate under imbalanced conditions. All combinations of class imbalance cases were tested in the evaluation. For small training datasets such as STL-10 and Caltech-256, we did not create an imbalanced condition. Under the class balanced condition, each experiment was repeated 10 times.
	
	\textbf{SVHN:} To test the transferability in the same domain, we set EMNIST as the source task and SVHN as the target task, where both are single-digit images. To reduce the image dimension gap, the SVHN images are converted to grayscale. For the imbalance 10:1 ratio set-up in training, the number of majority class images is 5000, and one of the minority class images is 500 in SVHN.
	
	\textbf{F-MNIST:} To test the transferability in the different domains, we set EMNIST as the source task and F-MNIST as the target task which has images of fashion and clothing items. For imbalanced 40:1 ratio set-up in training, the number of majority class images is 6000 and that of minority class images is 150 in F-MNIST. 
	
	\textbf{STL-10:} STL-10 images are resized to 32 $\times$ 32 to match the image dimension of CIFAR-10. In this case, class imbalance is not applied, but a small training dataset where 500 images per class are used in STL-10. 
	
	\textbf{CINIC-10:} CINIC-10 dataset contains 9000 training images in each class and we sampled 900 images from each minor class for imbalance 10:1 ratio.
	
	\textbf{Caltech-256:} Caltech 256 contains 257 object categories and 30607 images. In this study, we sampled 60 training samples for each category, referring to \cite{li2018explicit, li2018delta}. We resized the input images to 256 $\times$ 256, followed by data augmentation operations of random mirror and random crop to 224 $\times$ 224.
	
	\textbf{Food-101:} Food-101 contains 101 food categories with 101000 images. A total of 750 training images and 250 test images were provided for each class. We sampled 250 training samples from minority classes for an imbalance of 5:1 ratio set-up during training. 
	
	All the experiments were implemented using Python scripts in the PyTorch framework and tested on a Linux system. Training procedures were performed on NVIDIA Tesla V100 GPUs. The training details can be found in \ref{networkarchitectures}.
	
	\subsection{Results and comparisons}
	
	\begin{table}[t]
		\centering
		\caption{Comparison results (in \%) of our method with competing algorithms. The numbers are represented as \textit{mean$\pm$std}. Four datasets are under the data-imbalanced condition, and two datasets are under the data-balanced conditions. The number of the training datasets is small.}
		\label{comparisonresults}
		\begin{tabular}{lccccc}
			\hline
			Dataset & \multirow{2}{*}{Original} & \multirow{2}{*}{Fine-tuning} & \multirow{2}{*}{DELTA} & DIFA + & \multirow{2}{*}{Ours(DFG)}\\
			(IR) & & & & cMWGAN & \\
			\hline
			SVHN & \multirow{2}{*}{76.57 $\pm$ 1.65} & \multirow{2}{*}{76.66 $\pm$ 0.65} & \multirow{2}{*}{78.47 $\pm$ 0.57} & \multirow{2}{*}{76.64 $\pm$ 1.32} & \multirow{2}{*}{\textbf{80.81 $\pm$ 0.25}}  \\
			(10:1) &   &  &  &  & \\
			F-MNIST & \multirow{2}{*}{77.13 $\pm$ 3.24} & \multirow{2}{*}{75.39$\pm$ 3.54} & \multirow{2}{*}{78.91 $\pm$ 2.27} & \multirow{2}{*}{78.27 $\pm$ 2.32} & \multirow{2}{*}{\textbf{82.65 $\pm$ 0.28}}\\
			(40:1) &   &  &  &  & \\
			STL-10 & \multirow{2}{*}{66.73 $\pm$ 0.67} & \multirow{2}{*}{72.89 $\pm$ 0.38} & \multirow{2}{*}{79.41 $\pm$ 0.27} & \multirow{2}{*}{80.08 $\pm$ 0.19} & \multirow{2}{*}{\textbf{81.09 $\pm$ 0.17}} \\
			(1:1) \\
			CINIC-10 & \multirow{2}{*}{58.14 $\pm$ 3.42} & \multirow{2}{*}{64.42 $\pm$ 0.81} & \multirow{2}{*}{68.89 $\pm$ 0.18} & \multirow{2}{*}{71.82 $\pm$ 0.35} & \multirow{2}{*}{\textbf{72.09 $\pm$ 0.20}} \\
			(10:1) \\
			Caltech-256 & \multirow{2}{*}{43.30 $\pm$ 0.34} & \multirow{2}{*}{81.99 $\pm$ 0.12} & \multirow{2}{*}{85.33 $\pm$ 0.15} & \multirow{2}{*}{82.23 $\pm$ 0.32} & \multirow{2}{*}{\textbf{86.29 $\pm$ 0.19}} \\
			(1:1) \\
			Food-101 & \multirow{2}{*}{30.17 $\pm$ 1.72} & \multirow{2}{*}{68.99 $\pm$ 0.42} & \multirow{2}{*}{72.03 $\pm$ 0.35} & \multirow{2}{*}{71.71 $\pm$ 0.08} & \multirow{2}{*}{\textbf{76.00 $\pm$ 0.36}} \\
			(5:1) \\
			\hline
		\end{tabular}
	\end{table}

	To establish a baseline for comparison, we denote ``Original'' as the performance of a classification model trained on the target dataset. ``Original'' is trained under the class-imbalanced condition in the case of SVHN, F-MNIST, CINIC-10, and Food-101. In the case of STL-10 and Caltech-256, ``Original'' is trained under the class-balanced condition that contains a small number of train data.
	
	As mentioned, ``Original'' in Table \ref{comparisonresults} is trained without transfer learning and shows accuracies tested on the target test dataset achieved with classifiers trained on the target training dataset. For the class imbalance problem, in which classes are selected as majority classes, is an important factor in determining the performance of the neural network, and the deviation of result accuracies is large. Thus, accuracies under the class imbalanced condition show the mean and standard deviation. 
	
	The proposed method is compared with previous fine-tuning methods.
	\begin{itemize}
		\item Fine-tuning: Traditional transfer learning method by fixing weights before the last convolution block and fine-tuning the rest of them.
		\item DELTA \cite{li2018delta}
		\item DIFA+cMWGAN: The adversarial feature augmentation method based on \cite{volpi2018adversarial}. We modified the structures of GAN with cMWGAN \cite{zhang2019feature}.
	\end{itemize}
	
	For a fair comparison with the other methods, we conducted experiments on the same neural network architecture for classification, such as LeNet5, VGGNet-16, and ResNet-50. Table \ref{comparisonresults} shows the results of the proposed DFG method compared to the fine-tuning, DELTA, and DIFA+cMWGAN methods. The results are the top-1 classification accuracies. The results in Table \ref{comparisonresults} show that our method based on DFG leads to accuracies higher than other methods for three different classifiers and six different benchmark datasets.
	
	\subsection{Qualitative results and ablation study}
	
\begin{figure}
		\centering
		\subfigure[]{
			\includegraphics[width=0.9\columnwidth]{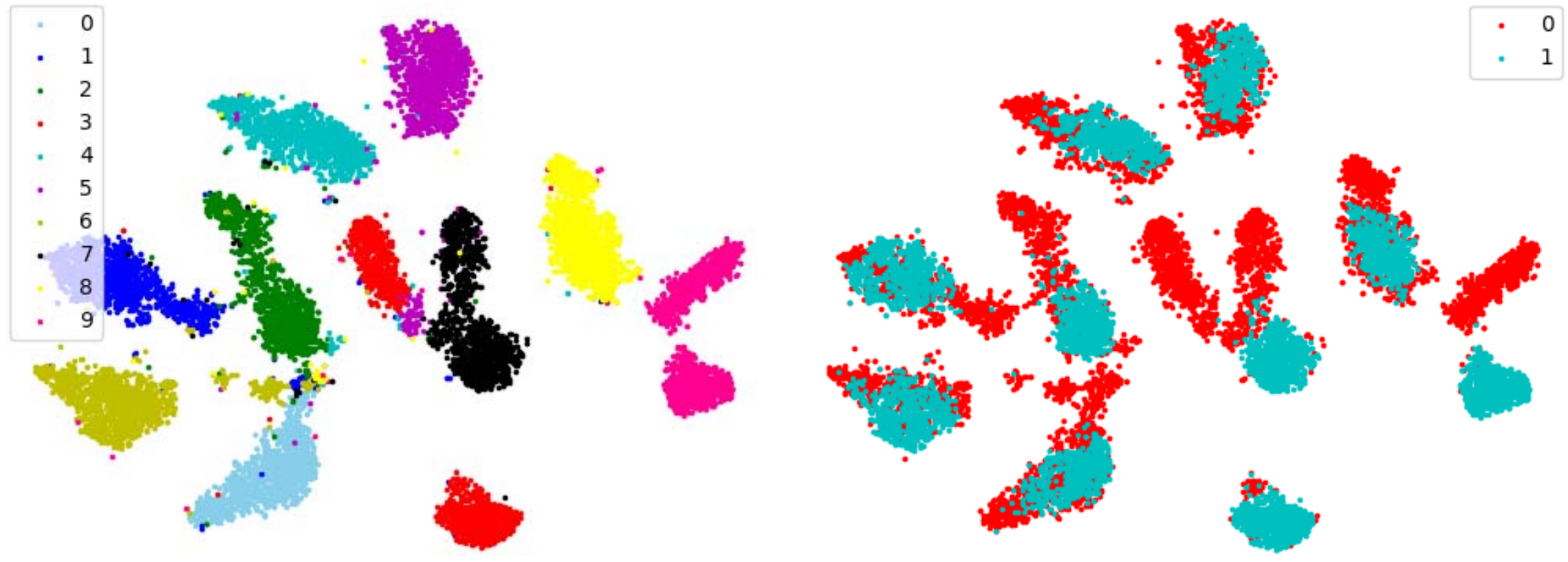}
		}
		\subfigure[]{
			\includegraphics[width=0.9\columnwidth]{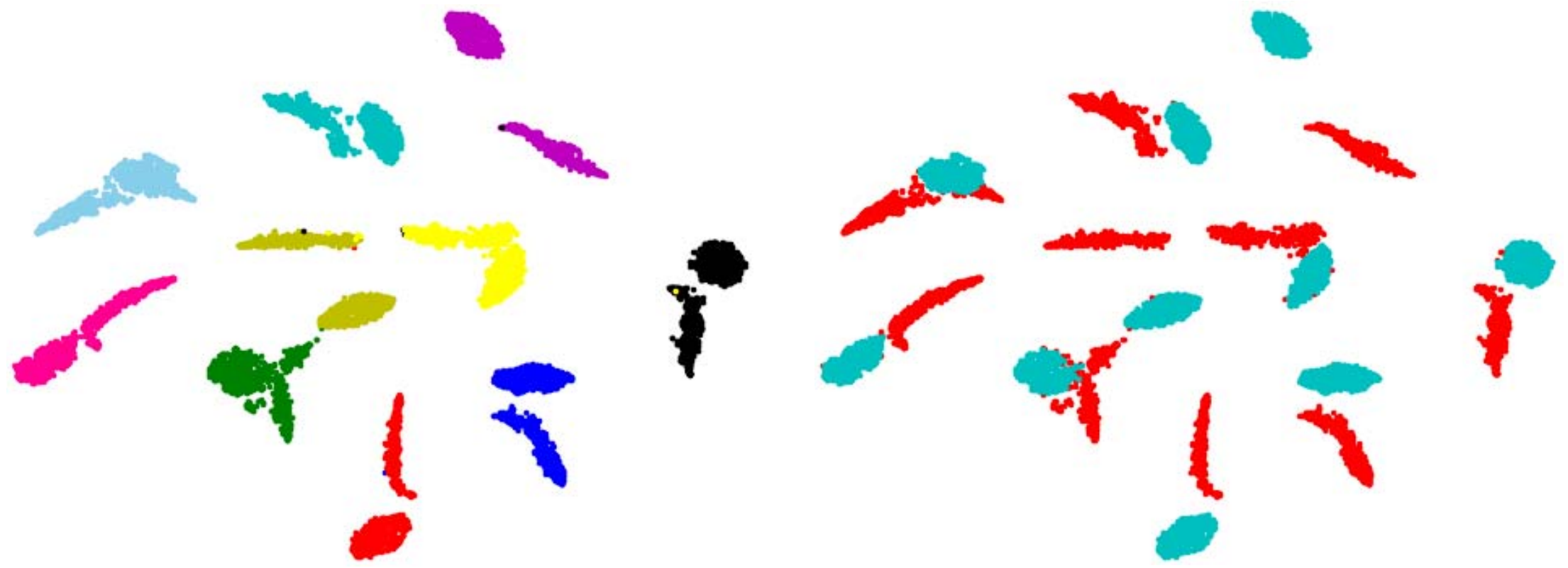}
		}
		\caption{t-SNE visualization of feature distribution on (a) SVHN and (b) STL. In the left part, different colors indicate different classes. In the right part, red and cyan dots indicate real and generated features, respectively.}
		\label{tsne}
	\end{figure}

	Although the proposed DFG method shows higher accuracy than the other methods, we would like to better understand the benefits brought by DFG and visualize how closely the proposed DFG generated the distribution of features resembled that of the original in two-dimensional spaces. We apply PCA \cite{wold1987principal} and t-SNE \cite{maaten2008visualizing} analyses of real and generated features. Figure \ref{tsne} shows comparisons between the real and generated features. Different colors indicate different classes in the left part, and two different colors indicate real and generated features in the right part. From a qualitative point of view, real and generated features are distributed similarly, and the inter-class structure is preserved. 
	
\begin{figure}
		\centering
		\subfigure[]{
			\includegraphics[width=0.18\linewidth]{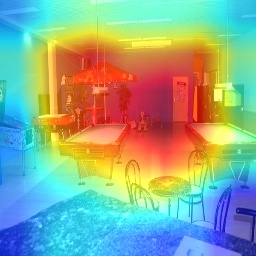}
		}
		\subfigure[]{
			\includegraphics[width=0.18\linewidth]{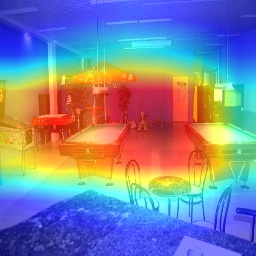}
		}
		\subfigure[]{
			\includegraphics[width=0.18\linewidth]{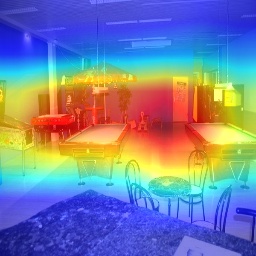}
		}
		\subfigure[]{
			\includegraphics[width=0.18\linewidth]{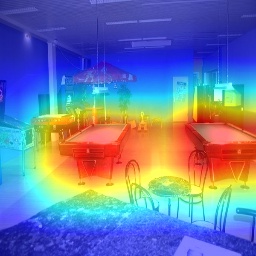}
		}
		\setcounter{subfigure}{0}
		\subfigure[]{
			\includegraphics[width=0.18\linewidth]{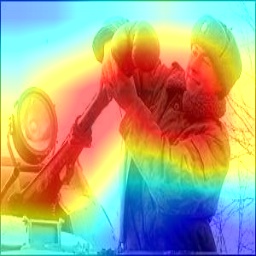}
		}
		\subfigure[]{
			\includegraphics[width=0.18\linewidth]{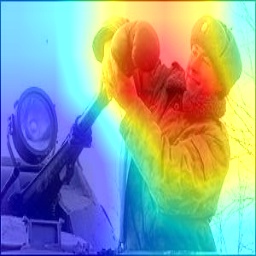}
		}
		\subfigure[]{
			\includegraphics[width=0.18\linewidth]{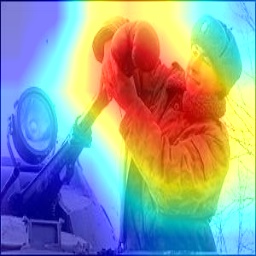}
		}
		\subfigure[]{
			\includegraphics[width=0.18\linewidth]{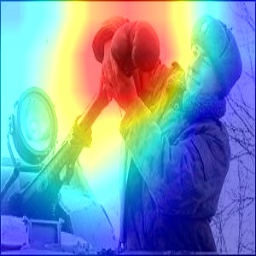}
		}
		\setcounter{subfigure}{0}
		\subfigure[]{
			\includegraphics[width=0.18\linewidth]{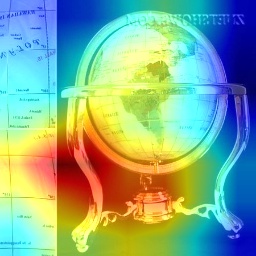}
		}
		\subfigure[]{
			\includegraphics[width=0.18\linewidth]{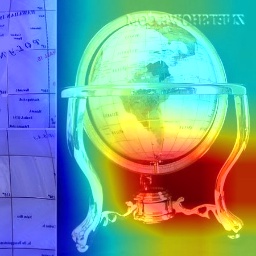}
		}
		\subfigure[]{
			\includegraphics[width=0.18\linewidth]{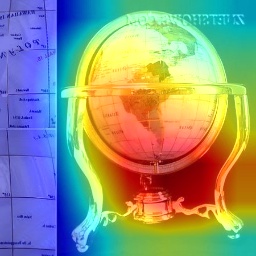}
		}
		\subfigure[]{
			\includegraphics[width=0.18\linewidth]{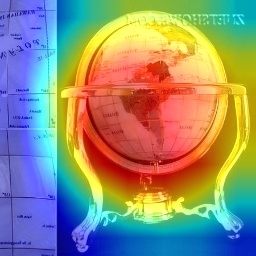}
		}
		\setcounter{subfigure}{0}
		\subfigure[]{
			\includegraphics[width=0.18\linewidth]{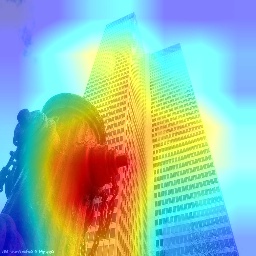}
		}
		\subfigure[]{
			\includegraphics[width=0.18\linewidth]{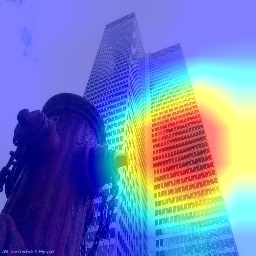}
		}
		\subfigure[]{
			\includegraphics[width=0.18\linewidth]{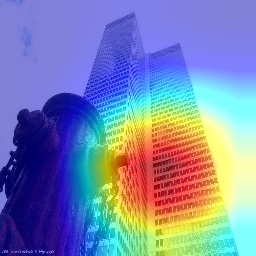}
		}
		\subfigure[]{
			\includegraphics[width=0.18\linewidth]{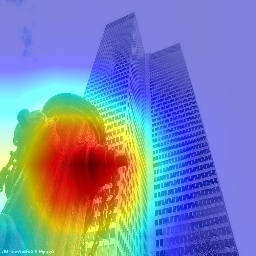}
		}
		\setcounter{subfigure}{0}
		\subfigure[]{
			\includegraphics[width=0.18\linewidth]{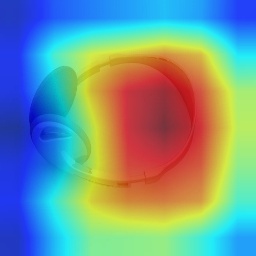}
		}
		\subfigure[]{
			\includegraphics[width=0.18\linewidth]{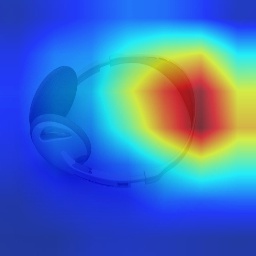}
		}
		\subfigure[]{
			\includegraphics[width=0.18\linewidth]{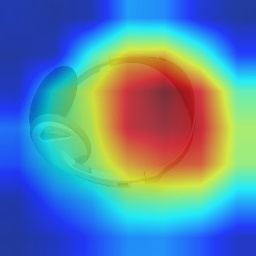}
		}
		\subfigure[]{
			\includegraphics[width=0.18\linewidth]{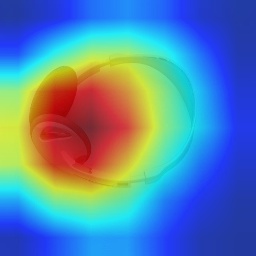}
		}
		\caption{Effect of the discriminative feature generation method on the Caltech-256 dataset: (a) original, (b) fine-tuning, (c) DELTA, and (d) DFG.}
		\label{cam_caltech}
	\end{figure}

\begin{figure}
		\centering
		\subfigure[]{
			\includegraphics[width=0.18\linewidth]{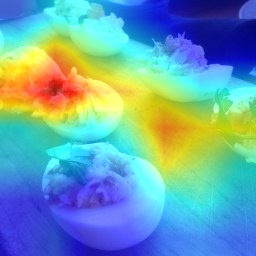}
		}
		\subfigure[]{
			\includegraphics[width=0.18\linewidth]{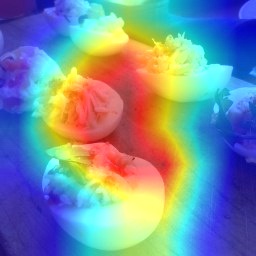}
		}
		\subfigure[]{
			\includegraphics[width=0.18\linewidth]{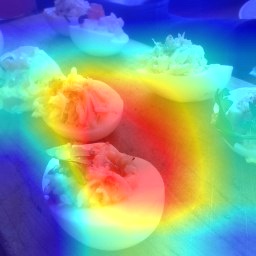}
		}
		\subfigure[]{
			\includegraphics[width=0.18\linewidth]{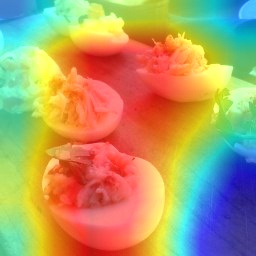}
		}
		\setcounter{subfigure}{0}
		\subfigure[]{
			\includegraphics[width=0.18\linewidth]{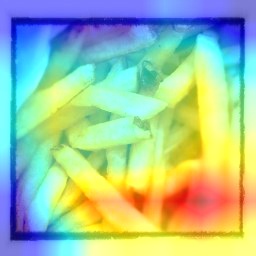}
		}
		\subfigure[]{
			\includegraphics[width=0.18\linewidth]{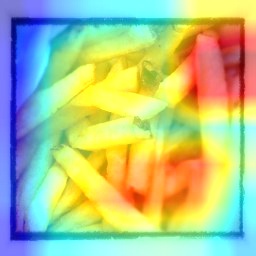}
		}
		\subfigure[]{
			\includegraphics[width=0.18\linewidth]{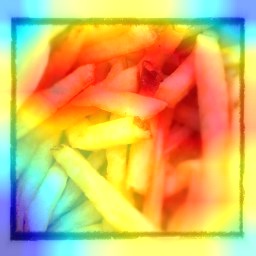}
		}
		\subfigure[]{
			\includegraphics[width=0.18\linewidth]{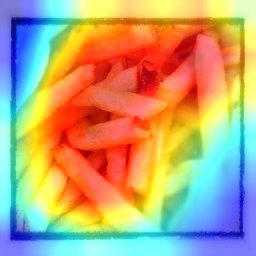}
		}
		\setcounter{subfigure}{0}
		\subfigure[]{
			\includegraphics[width=0.18\linewidth]{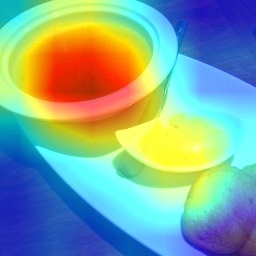}
		}
		\subfigure[]{
			\includegraphics[width=0.18\linewidth]{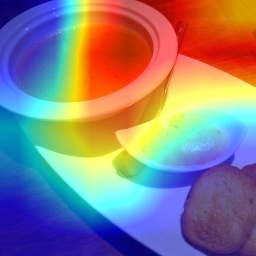}
		}
		\subfigure[]{
			\includegraphics[width=0.18\linewidth]{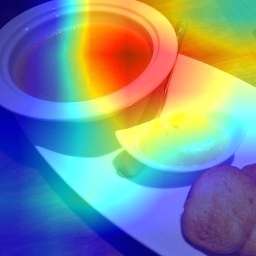}
		}
		\subfigure[]{
			\includegraphics[width=0.18\linewidth]{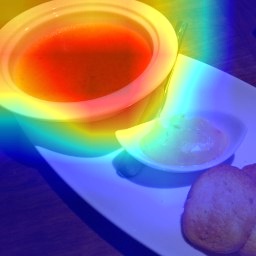}
		}
		\setcounter{subfigure}{0}
		\subfigure[]{
			\includegraphics[width=0.18\linewidth]{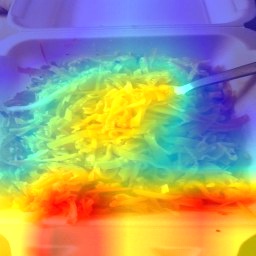}
		}
		\subfigure[]{
			\includegraphics[width=0.18\linewidth]{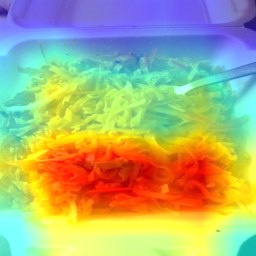}
		}
		\subfigure[]{
			\includegraphics[width=0.18\linewidth]{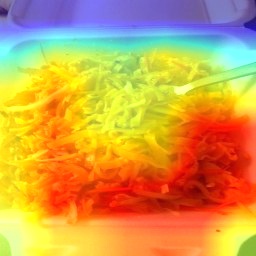}
		}
		\subfigure[]{
			\includegraphics[width=0.18\linewidth]{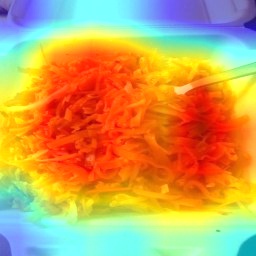}
		}
		\setcounter{subfigure}{0}
		\subfigure[]{
			\includegraphics[width=0.18\linewidth]{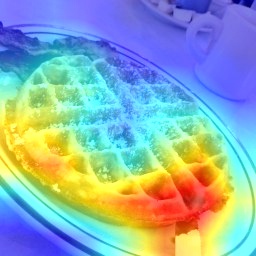}
		}
		\subfigure[]{
			\includegraphics[width=0.18\linewidth]{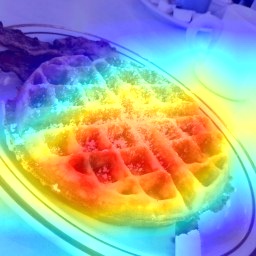}
		}
		\subfigure[]{
			\includegraphics[width=0.18\linewidth]{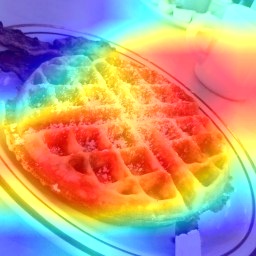}
		}
		\subfigure[]{
			\includegraphics[width=0.18\linewidth]{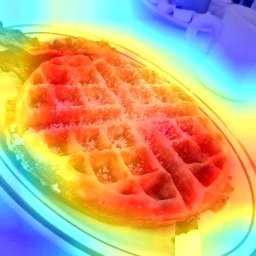}
		}
		\caption{Effect of the discriminative feature generation method on the Food-101 dataset: (a) original, (b) fine-tuning, (c) DELTA, and (d) DFG.}
		\label{cam_food101}
	\end{figure}

	Additionally, we analyzed the activation maps from the proposed DFG method using class activation maps \cite{zhou2016learning}. From Figures \ref{cam_caltech} and \ref{cam_food101}, we observe that the regions of the target object have higher activation than that of the original classifier and fine-tuning. Compared with the activation maps from the DELTA, the proposed method improved the concentration and activation degrees. 
	
	
	\begin{figure}
		\centering
		\includegraphics[width=0.9\columnwidth]{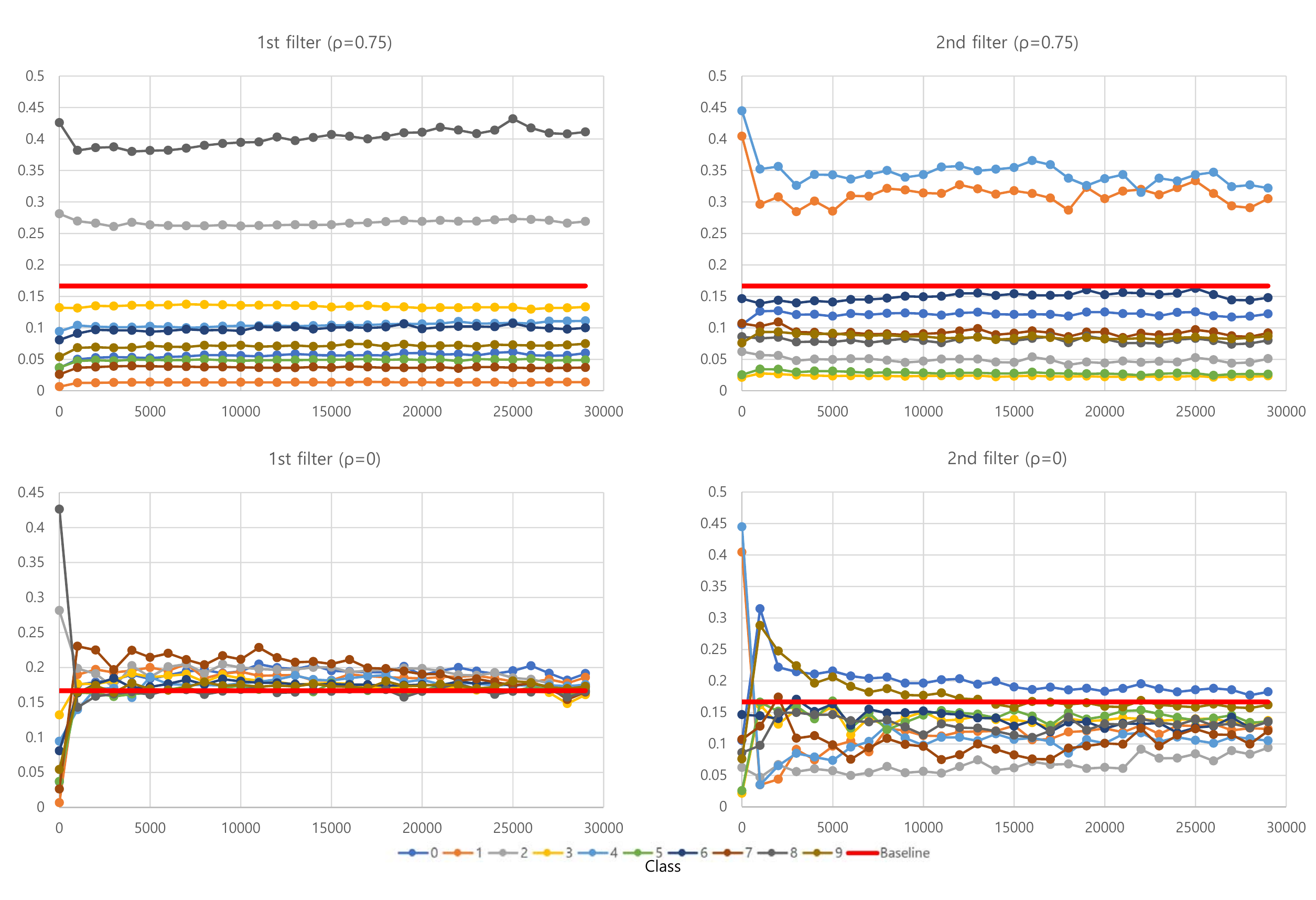}
		\caption{Distribution of the weights of two representative filters for the SVHN dataset. The red straight line is a baseline that indicates the equality of all filters.}
		\label{filterweight}
	\end{figure}
	
	We carried out an ablation study to better understand how the filter weight changes during the training procedure. We perform an experiment by changing the $\rho$ value in (\ref{eq:classwisegenerator_attention_weightedsum}). Figure \ref{filterweight} presents the distribution of the weights of two representative filters in the case of the SVHN dataset for two different hyperparameters: $\rho = 0$ and $0.75$. The first row in Figure \ref{filterweight} represents the distribution of the weights of two representative filters with $\rho=0.75$, and the second row represents the distribution of the weights of the same filters with $\rho=0$. The red straight line is the line of equality for all filters. For instance, if a weight value is placed above the line of equality, it means that the importance of the filter for the class is higher than the average of all filters. The case with $\rho=0$ involves the training of the feature generator with the weights of the filters by updating from the feature extractor trained on the target dataset, and the case with $\rho=0.75$ involves the training of the feature generator with the weights of the filters by two feature extractors: one is trained on the source dataset and the other is trained on the target dataset in the training procedure. We observed a significant difference between the two distributions of the weights. The distribution of the weights of the filters with $\rho=0$ converges to the equality baseline, whereas the distribution of the weights of the filters with $\rho=0.75$ maintains the characteristics of the weight for each class label. Our hypothesis of a possible cause of the difference is that the feature extractor trained on the target dataset tends to be trained to activate all filters and be overfitted to the target dataset. The feature extractor trained on the source dataset inactivates several convolution filters, and a small number of filters are activated. The average accuracy with $\rho=0.75$ is 80.81\% and that with $\rho=0$ is 79.98\%. The weights of the filters with $\rho=0.75$, using the weight values from the feature extractor trained on the source dataset, prevent overfitting and help to generate discriminative features for each class label.	
	\begin{figure}
		\centering
		\includegraphics[width=0.8\columnwidth]{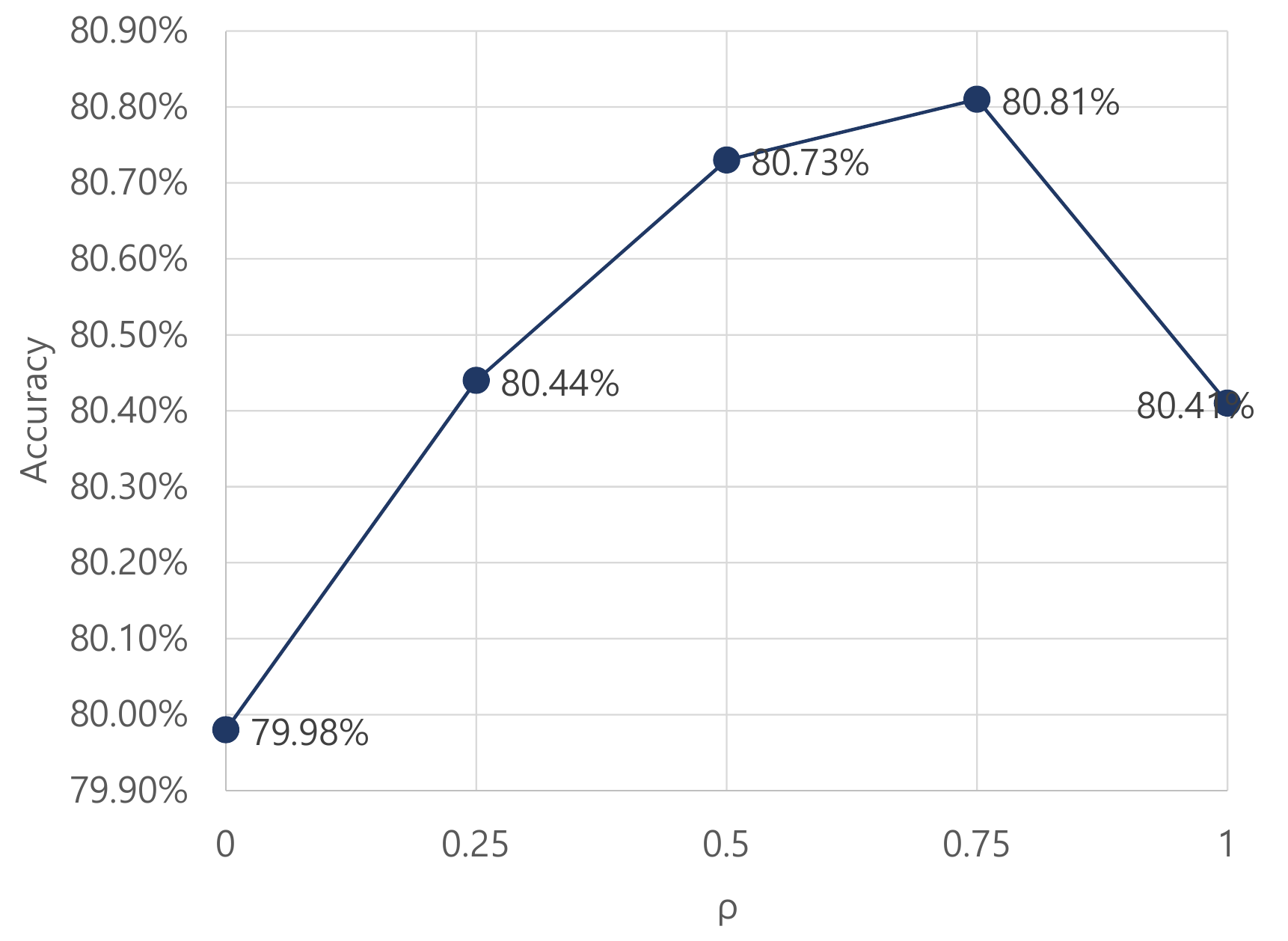}
		\caption{Classification accuracy on SVHN using DFG by changing hyperparameter $\rho$.}
		\label{accuracygamma}
	\end{figure}
	
	Figure \ref{accuracygamma} shows the effect of hyperparameter $\rho$ on the classification accuracy on the SVHN dataset using DFG. In this case, DFG with hyperparameter $\rho=0.75$ outperformed that with other hyperparameters including the cases that the weights of the filters by the feature extractor only trained on the source or the target dataset. The test accuracy varies smoothly according to hyperparameter $\rho$, and combination of the two different weights of a filter yields a better accuracy. 
	
	\section{Conclusions}
	\label{conclusions}
	
	In this paper, we proposed a novel DFG method using attention maps in the feature space. The structure of the feature classifier network of the independent network was employed in the proposed GAN structure. To generate discriminative features, we adopted supervised attention mechanisms for each class label and applied the mechanisms to the feature generator network for the filter weights and feature activation. The experimental results showed that the proposed DFG method improved the accuracy of the classification models of LeNet-5, VGG-16, and ResNet-50 by 4.24\%, 5.52\%, 14.73\%, 13.95\%, 42.99\%, and 45.83\% on six different benchmark datasets, respectively. Comparing to the transfer learning method, DELTA, and the other adversarial feature augmentation methods, the proposed DFG method provided the largest improvement in classification performance. Furthermore, we visualized the distribution of generated features by applying PCA and t-SNE and analyzed the activation maps from the proposed method using class activation maps. The qualitative results showed that the features generated by the proposed method are distributed similarly to the real features and the proposed method improved the concentration and activation degrees in the class activation maps.
	
	In the future, we will extend the proposed method to real-world problems and contemplate further accuracy improvement in the feature generation method. Although the proposed method improved the performance of the classification under the class-imbalanced conditions, we believe there is room for improvement as a gap still exists between our method and the balanced condition.

	\section*{Acknowledgments}
	This work was supported by the KIST Europe Institutional Program (Project No. 12020).

	\bibliography{mybibfile}
	
	\appendix
	\section{Network Architectures and Implementation Details}
	\label{networkarchitectures}
	
	We implemented the proposed DFG method with LeNet-5 \cite{lecun1998gradient}, VGGNet-16 \cite{simonyan2014very}, and ResNet-50 \cite{he2016deep} as the feature extractor and feature classifier. We divided the original architecture of the classifiers into the feature extractor and feature classifier. The architecture of the feature extractor and feature classifier is presented in Tables \ref{tab:LeNet}, \ref{tab:VGGNet}, and \ref{tab:ResNet}. The output dimensions of the feature extractor and the feature generator should be the same. The architecture of our feature discriminator and feature generator closely follow the DCGAN architecture model \cite{radford2015unsupervised}, which is an extended model of the GAN that uses 3--4 transpose convolution layers in the generator and three convolution layers in the feature discriminator. The architecture of the generator and the discriminator for LeNet-5, VGGNet-16, and ResNet-50 are shown in Tables \ref{tab:dcgan_lenet5}, \ref{tab:dcgan_vgg16}, \ref{tab:dcgan_resnet50}, respectively.
	
	We used extended MNIST digits (EMNIST) \cite{cohen2017emnist}, CIFAR-10 \cite{krizhevsky2009learning}, and ImageNet \cite{deng2009imagenet} as the source domain for LeNet-5, VGG-16, and ResNet-50, respectively. For transfer learning, we followed the same procedure as in \cite{li2018delta} because of the close relationship between the behavior regularization in this study and that one. After adopting the pre-trained weight only for ResNet-50 and before fine-tuning the network with the target dataset, we replaced the last layer of the base network with random initialization in a suit for the target dataset because the number of classes in the source dataset, ImageNet, is different from those in the target datasets, Caltech-256 \cite{griffin2007caltech} and Food-101 \cite{bossard2014food}.

	\textbf{LeNet-5:} LeNet-5 model was used for SVHN \cite{netzer2011reading} and Fashion-MNIST (F-MNIST) \cite{xiao2017fashion}. To remove the image dimension difference, SVHN images were converted to grayscale and F-MNIST images were resized to 32 $\times$ 32 with zero-padding. The input images were normalized to zero-mean and the pixel values to range $[-1, 1]$. We did not use any data augmentation method for preprocessing. The weights of the feature extractor and feature classifier were pre-trained using EMNIST. The weights of the feature generator and feature discriminator were initialized using a Xavier procedure, and the biases were set to zero \cite{glorot2010understanding}. We used a batch size of 64 and the Adam optimizer for all networks with parameters $\beta_1=0.5$ and $\beta_2=0.9$ \cite{kingma2014adam}. The learning rates of the feature extractor, feature classifier, feature generator, and feature discriminator for SVHN were $2.0\times 10^{-5}$, $2.0\times 10^{-5}$, $2.0\times 10^{-4}$, and $1.0\times 10^{-4}$, respectively ($\eta_E$, $\eta_C$, $\eta_G$, and $\eta_D$ in Algorithm \ref{Algopseudocode}.). The learning rates for F-MNIST were $\eta_E=\eta_C=1.0\times 10^{-5}$ and $\eta_G=\eta_D=1.0\times 10^{-4}$. The training steps were 20000 iterations. The parameters in Algorithm \ref{Algopseudocode} were set to $n_{C1}=2$, $n_{C2}=10$, $n_W=5000$, $\rho=0.75$, $\delta=0.95$, and $\alpha=\beta=\gamma=0.333$.
	
	\textbf{VGGNet-16:} VGGNet-16 model was used for STL-10 \cite{coates2011analysis} and CINIC-10 \cite{darlow2018cinic}. CIFAR-10 (similar class labels to STL-10 and CINIC-10) was used for pre-training as the source domain. We downscaled the 96 $\times$ 96 image dimension of STL-10 to match the 32 $\times$ 32 dimension of CIFAR-10 and CINIC-10. The input images were normalized to zero-mean and the pixel values to range $[-1, 1]$. We applied data augmentation operations of random horizontal flip and random crop with 4-pixels padding. The weights of the feature extractor and feature classifier were pre-trained with CIFAR-10. The weights of the feature generator and the feature discriminator were initialized using a He procedure, and the biases were set to zero \cite{He_2015_ICCV}. We used a batch size of 64 and the Adam optimizer for all networks with parameters $\beta_1=0.5$ and $\beta_2=0.9$. The learning rates for STL-10 were set to $\eta_E=\eta_C=5.0\times 10^{-5}$ and $\eta_G=\eta_D=1.0\times 10^{-4}$. The learning rates for CINIC-10 were set to $\eta_E=\eta_C=2.0\times 10^{-5}$ and $\eta_G=\eta_D=1.0\times 10^{-4}$. The training steps for STL-10 and CINIC-10 were 20000 and 40000 iterations, respectively. The parameters in Algorithm \ref{Algopseudocode} were set to $n_{C1}=2$, $n_{C2}=10$, $n_W=5000$, $\rho=0.75$, $\delta=0.95$, and $\alpha=\beta=\gamma=0.333$.
	
	\textbf{ResNet-50:} ResNet-50 model was used for Caltech-256 \cite{griffin2007caltech} and Food-101 \cite{bossard2014food}. ImageNet was used as the source domain for pre-training. We replaced the last fully connected layer of ResNet-50 because of the different number of classes in ImageNet, Caltech-256, and Food-101 (1000, 257, and 101, respectively). For ResNet-50, the input images were normalized to zero-mean and the pixel values to range $[-1, 1]$ and resized to 256 $\times$ 256. We applied the data augmentation operation of the random crop to 224 $\times$ 224 and random horizontal flip. For the Caltech-256 dataset, we sampled 60 training samples and 20 testing samples for each category, according to \cite{li2018explicit, li2018delta}. The weights of the feature extractor and the feature classifier were initialized by the model pre-trained on ImageNet, provided by Torchvision \cite{Marcel2010torchvision}, and the feature generator and the feature discriminator were initialized using a He initialization, and the biases were set to zero. We used a batch size of 64 and the Adam optimizer with parameters $\beta_1=0.5$ and $\beta_2=0.9$ for the feature generator and the feature discriminator, and stochastic gradient descent with a momentum of 0.9 for optimizing the feature extractor and the feature classifier. The learning rate for the feature generator and the feature discriminator was set to $\eta_G=\eta_D=1.0\times 10^{-4}$, and the learning rate for the feature extractor and feature classifier started with 0.01 and was divided by 10 after 20000 iterations. The training steps were completed at 40000 iterations. The parameters in Algorithm \ref{Algopseudocode} were set to $n_{C1}=1$, $n_{C2}=5$, $n_W=2000$, $\rho=0.75$, $\delta=0.95$, and $\alpha=\beta=\gamma=0.333$.

	Additionally, we used the following publicly available source code to implement our benchmarks.
	\begin{itemize}
		\item DELTA \cite{li2018delta}: \url{https://github.com/lixingjian/DELTA}
		\item DIFA+cMWGAN \cite{volpi2018adversarial, zhang2019feature}: \\ \url{https://github.com/ricvolpi/adversarial-feature-augmentation}
	\end{itemize}

	\begin{table}
		\centering
		\caption{Architecture of feature extractor and feature classifier based on LeNet-5 used for SVHN and fashion-MNIST datasets.}
		\label{tab:LeNet}		
		\begin{tabular}{llcc}
			\hline
			\textbf{Network} & \textbf{Layers} & \textbf{Act. Func.} & \textbf{Dimension} \\
			\hline
			\multirow{3}{*}{Feature Extractor} & Input Image & - & 32 $\times$ 32 $\times$ 1\\
			& Conv 5 $\times$ 5 & ReLU & 28 $\times$ 28 $\times$ 6\\
			& Max Pooling 2 $\times$ 2 & - & 14 $\times$ 14 $\times$ 6\\
			\hline
			\multirow{7}{*}{Feature Classifier} & Input Feature & - & 14 $\times$ 14 $\times$ 6\\
			&Conv 5 $\times$ 5 & ReLU & 10 $\times$ 10 $\times$ 16\\
			&Max Pooling 2 $\times$ 2 & - & 5 $\times$ 5 $\times$ 16\\
			&Fully Connected & ReLU & 120\\
			&Fully Connected & ReLU & 84\\
			&Fully Connected & - & 10\\
			&SoftMax & - & 10\\
			\hline
		\end{tabular}
	\end{table}
	
	\begin{table}
		\caption{Architecture of the generator and the discriminator for LeNet-5 classifier on SVHN and fashion-MNIST datasets.}
		\label{tab:dcgan_lenet5}
		\centering
		\begin{tabular}{llcc}
			\hline
			\textbf{Network} & \textbf{Layers} & \textbf{Act. func.} & \textbf{Dimension} \\
			\hline
			\multirow{9}{*}{Feature generator} & Input & - & 110\\
			& FC & - & 4 $\times$ 4 $\times$ 128\\
			& BatchNorm & - & 4 $\times$ 4 $\times$ 128\\
			& DeConv 3 $\times$ 3, st = 2 & lReLU & 9 $\times$ 9 $\times$ 48\\
			& BatchNorm & - & 9 $\times$ 9 $\times$ 48\\
			& DeConv 3 $\times$ 3, st = 1 & lReLU & 11 $\times$ 11 $\times$ 12\\
			& BatchNorm & - & 11 $\times$ 11 $\times$ 12\\
			& DeConv 4 $\times$ 4, st = 1 & Tanh & 14 $\times$ 14 $\times$ 6\\
			& BatchNorm & - & 14 $\times$ 14 $\times$ 6\\
			\hline
			\multirow{5}{*}{Feature discriminator} & Input & - & 14 $\times$ 14 $\times$ 6\\
			& Conv 5 $\times$ 5, st = 2 & lReLU & 7 $\times$ 7 $\times$ 32\\
			& Conv 5 $\times$ 5, st = 2 & lReLU & 4 $\times$ 4 $\times$ 64\\
			& Conv 5 $\times$ 5, st = 2 & lReLU & 2 $\times$ 2 $\times$ 128\\
			& FC & - & 512\\ 
			\hline
		\end{tabular}
	\end{table}
	
	\begin{table}
		\caption{Architecture of feature extractor and feature classifier based on VGGNet-16 used for STL-10 and CINIC-10 datasets.}
		\label{tab:VGGNet}
		\centering
		\begin{tabular}{llcc}
			\hline
			\textbf{Network} & \textbf{Layers} & \textbf{Act. Func.} & \textbf{Dimension} \\
			\hline
			\multirow{3}{*}{Feature Extractor} & Input Image & - & 32 $\times$ 32 $\times$ 3\\
			& 2$\times$Conv 3 $\times$ 3 & ReLU & 32 $\times$ 32 $\times$ 64\\
			& Max Pooling 2 $\times$ 2, st = 2 & - & 16 $\times$ 16 $\times$ 64\\
			\hline
			\multirow{14}{*}{Feature Classifier} & Input Feature & - & 16 $\times$ 16 $\times$ 64\\
			& 2$\times$Conv 3 $\times$ 3 & ReLU & 16 $\times$ 16 $\times$128\\
			& Max Pooling 2 $\times$ 2, st = 2 & - & 8 $\times$ 8 $\times$ 128\\
			& 3$\times$Conv 3 $\times$ 3 & ReLU & 8 $\times$8 $\times$ 256\\
			& Max Pooling 2 $\times$ 2, st = 2 & - & 4 $\times$ 4 $\times$ 256\\
			& 3$\times$Conv 3 $\times$ 3 & ReLU & 4 $\times$ 4 $\times$ 512\\
			& Max Pooling 2 $\times$ 2, st = 2 & - & 2 $\times$ 2 $\times$ 512\\
			& 3$\times$Conv 3 $\times$ 3 & ReLU & 2 $\times$ 2 $\times$ 512\\
			& Max Pooling 2 $\times$ 2, st = 2 & - & 1 $\times$ 1 $\times$ 512\\
			& Adaptive Avg Pooling & - & 7 $\times$ 7 $\times$ 512\\
			& Fully Connected & ReLU & 4096\\
			& Fully Connected & ReLU & 512\\
			& Fully Connected & - & 10\\
			& SoftMax & - & 10\\
			\hline
		\end{tabular}
	\end{table}
	
	\begin{table}
		\caption{Architecture of the generator and the discriminator for VGGNet-16 classifier on STL-10 and CINIC-10 datasets.}
		\label{tab:dcgan_vgg16}
		\centering
		\begin{tabular}{llcc}
			\hline
			\textbf{Network} & \textbf{Layers} & \textbf{Act. func.} & \textbf{Dimension} \\
			\hline
			\multirow{11}{*}{Feature generator} & Input & - & 110\\
			& FC & - & 4 $\times$ 4 $\times$ 128\\
			& BatchNorm & - & 4 $\times$ 4 $\times$ 128\\
			& DeConv 4 $\times$ 4, st = 2 & lReLU & 8 $\times$ 8 $\times$ 128\\
			& BatchNorm & - & 8 $\times$ 8 $\times$ 128\\
			& DeConv 4 $\times$ 4, st = 2 & lReLU & 16 $\times$ 16 $\times$ 128\\
			& BatchNorm & - & 16 $\times$ 16 $\times$ 128\\
			& DeConv 4 $\times$ 4, st = 1 & lReLU & 17 $\times$ 17 $\times$ 64\\
			& BatchNorm & - & 17 $\times$ 17 $\times$ 64\\
			& DeConv 4 $\times$ 4, st = 1 & Tanh & 16 $\times$ 16 $\times$ 64\\
			& BatchNorm & - & 16 $\times$ 16 $\times$ 64\\
			\hline
			\multirow{5}{*}{Feature discriminator} & Input & - & 16 $\times$ 16 $\times$ 64\\
			& Conv 5 $\times$ 5, st = 2 & lReLU & 8 $\times$ 8 $\times$ 32\\
			& Conv 5 $\times$ 5, st = 2 & lReLU & 4 $\times$ 4 $\times$ 64\\
			& Conv 5 $\times$ 5, st = 2 & lReLU & 2 $\times$ 2 $\times$ 128\\
			& FC & - & 512\\ 
			\hline
		\end{tabular}
	\end{table}
	
	\begin{table}
		\caption{Architecture of feature extractor and feature classifier based on ResNet-50 used for Caltech-256 and Food-101 datasets.}
		\label{tab:ResNet}
		\centering
		\begin{tabular}{llcc}
			\hline
			\textbf{Network} & \textbf{Layers} & \textbf{Act. Func.} & \textbf{Dimension} \\
			\hline
			\multirow{5}{*}{Feature Extractor} & Input Image & - & 224 $\times$ 224 $\times$ 3\\
			& Conv 7 $\times$ 7 st = 2 & ReLU & 112 $\times$ 112 $\times$ 64\\
			& BatchNorm & - & 112 $\times$ 112 $\times$ 64\\
			& Max Pooling 3 $\times$ 3, st = 2 & - & 55 $\times$ 55 $\times$ 64\\
			& Residual Block \#1 & ReLU & 55 $\times$ 55 $\times$ 256\\
			\hline
			\multirow{7}{*}{Feature Classifier} & Input Feature & - & 55 $\times$ 55 $\times$ 256\\
			& Residual Block \#2 & ReLU & 28 $\times$ 28 $\times$ 512\\
			& Residual Block \#3 & ReLU & 14 $\times$ 14 $\times$ 1024\\
			& Residual Block \#4 & ReLU & 7 $\times$ 7 $\times$ 2048\\
			& Adaptive Avg Pooling & - & 1 $\times$ 1 $\times$ 2048\\
			& Fully Connected & - & No.Class\\
			& SoftMax & - & No.Class\\
			\hline
		\end{tabular}
	\end{table}
	
	\begin{table}
		\caption{Architecture of the generator and the discriminator for ResNet-50 classifier on Caltech-256 and Food-101 datasets.}
		\label{tab:dcgan_resnet50}
		\centering
		\begin{tabular}{llcc}
			\hline
			\textbf{Network} & \textbf{Layers} & \textbf{Act. func.} & \textbf{Dimension} \\
			\hline
			\multirow{11}{*}{Feature generator} & Input & - & 4900 + No.Class\\
			& FC & - & 4 $\times$ 4 $\times$ 512\\
			& BatchNorm & - & 4 $\times$ 4 $\times$ 512\\
			& DeConv 4 $\times$ 4, st = 2 & lReLU & 8 $\times$ 8 $\times$ 512\\
			& BatchNorm & - & 8 $\times$ 8 $\times$ 512\\
			& DeConv 4 $\times$ 4, st = 2 & lReLU & 16 $\times$ 16 $\times$ 256\\
			& BatchNorm & - & 16 $\times$ 16 $\times$ 256\\
			& DeConv 2 $\times$ 2, st = 2 & lReLU & 30 $\times$ 30 $\times$ 256\\
			& BatchNorm & - & 30 $\times$ 30 $\times$ 256\\
			& DeConv 2 $\times$ 2, st = 2 & Tanh & 56 $\times$ 56 $\times$ 256\\
			& BatchNorm & - & 56 $\times$ 56 $\times$ 256\\
			\hline
			\multirow{6}{*}{Feature discriminator} & Input & - & 56 $\times$ 56 $\times$ 256\\
			& Conv 5 $\times$ 5, st = 2 & lReLU & 28 $\times$ 28 $\times$ 32\\
			& Conv 5 $\times$ 5, st = 2 & lReLU & 14 $\times$ 14 $\times$ 64\\
			& Conv 5 $\times$ 5, st = 2 & lReLU & 7 $\times$ 7 $\times$ 128\\
			& FC & - & 6272\\ 
			& FC & - & 512\\ 
			\hline
		\end{tabular}
	\end{table}
	
\end{document}